\documentclass{article}

\usepackage[final]{corl_2020} 

\usepackage{graphicx}
\usepackage{enumitem}
\usepackage{makecell}
\usepackage{tabularx}
\usepackage{amsmath,amsfonts,amssymb}
\usepackage{xspace}
\usepackage[subtle,tracking=normal]{savetrees}
\usepackage{booktabs}
\usepackage{standalone}
\usepackage{caption}
\usepackage{pdfpages}

\title{Learning rich touch representations \\ through cross-modal self-supervision}

\author{
  Martina Zambelli\\
  Deepmind,
   London,
   UK \\
  \texttt{zambellim@google.com} \\
   \And
   Yusuf Aytar \\
   Deepmind,
   London,
   UK \\
   \texttt{yusufaytar@google.com} \\
   \And
   Francesco Visin \\
   Deepmind,
   London,
   UK \\
   \texttt{visin@google.com} \\
   \And
   Yuxiang Zhou \\
   Deepmind,
   London,
   UK \\
   \texttt{yuxiangzhou@google.com} \\
  \And
  Raia Hadsell \\
  Deepmind,
   London,
   UK \\
  \texttt{raia@google.com} \\
}

\begin{document}
\maketitle


\vspace*{-0.5cm}
\begin{abstract}
The sense of touch is fundamental in several manipulation tasks, but rarely used in robot manipulation. 
In this work we tackle the problem of learning rich touch features from cross-modal self-supervision. We evaluate them identifying objects and their properties in a few-shot classification setting.
Two new datasets are introduced using a simulated anthropomorphic robotic hand equipped with tactile sensors on both synthetic and daily life objects. 
Several self-supervised learning methods are benchmarked on these datasets, by evaluating few-shot classification on unseen objects and poses.
Our experiments indicate that cross-modal self-supervision effectively improves touch representation, and in turn has great potential to enhance robot manipulation skills. 
\end{abstract}
\vspace*{-.3cm}

\vspace*{-.28cm}
\section{Introduction}
\label{sec:introduction}
\vspace*{-.2cm}

Deep learning brought us many success stories in a wide range of fields within artificial intelligence, including vision, sound, and text understanding. One paradigm shift that became pervasive recently is that of learning rich high-level feature representations, and then applying them to a wide variety of downstream tasks. These rich representations are learned through supervised pretraining on large scale annotated datasets, such as ImageNet~\citep{deng2009imagenet,krizhevsky2012imagenet}, and then applied successfully to learn many downstream tasks with smaller amounts of annotated data, such as object detection~\citep{girshick2015region}, segmentation~\citep{long2015fully} and classification of new categories~\citep{donahue2014decaf}. This process, termed as \emph{finetuning}, became the main driving force of many new applications of deep learning where annotated data is only scarcely available. 

Most recently, the self-supervised learning approaches, which aim to learn rich representations from uncurated large scale data, removed the costly burden of annotations and enabled learning rich high-level representations in a large range of domains. In the image understanding domain self-supervised representations already reached the quality of representations learned through large scale annotated datasets~\citep{grill2020bootstrap,chen2020simple}. Self-supervised pretraining became the de-facto standard in text understanding~\citep{radford2018improving,devlin2018bert,yang2019xlnet} and even modalities like sound greatly benefited from rich representations learned through self-supervised approaches utilizing multiple modalities~\citep{arandjelovic2017look,alayrac2020self,korbar2018cooperative}. This methodology of obtaining rich features that enable \emph{sample-efficient} (i.e. few-shot) learning strongly resonates with us, as humans mostly learn new tasks in a similar manner even with very small amounts of data or experiences. In this paper we propose to learn rich touch representations that enable sample-efficient learning for touch-based perception (i.e. perception in the dark). We achieve this through cross-modal self-supervised learning by harnessing the natural correlation between vision and touch signals. 

The sense of touch is ubiquitous in our everyday life. While vision is possibly the predominant sensory modality through which we learn about and interact with our surroundings, touch plays a critical role in manipulation tasks. Indeed, it provides us with very heterogeneous and rich data, with direct measurements of ongoing contact forces during object interactions that allow us to infer friction, compliance, mass, and other physical properties of surfaces and objects. 
Critically, the mental models built combining data from vision, touch and possibly other modalities are very effectively exploited when vision is not available.

As a motivating example, consider a very common task such as looking for a set of keys in a bag. While we typically have a clear picture of how the keys look like, it is often hard to \emph{look} properly inside a bag, and we have to resort to other senses. In this situation, reaching with a hand in the bag allows us to explore using touch alone but, critically, exploiting the mental model previously built using data from the richer vision modality. 
This example is representative of a broader set of tasks where visual clues are limited or precluded when the task is performed, but are usually accessible otherwise. Consider for instance shifting gears while keeping the eyes on the road, grasping a glass of water in the dark or tying shoelaces without watching. 
While these tasks are relying mainly, if not only, on touch, they exploit a mental model of the objects that has been built throughout the person's life using all senses.

Such tasks are very common in our everyday life, and critically also in robot manipulation. Here, occlusions, restrictions of the visual field, or limited vision capabilities, can be a real impediment for the successful completion of many tasks. For instance understanding the identity and orientation of grasped objects when occluded by a robot hand can be challenging relying only on vision, but becomes much easier when paired with touch and proprioception.
Indeed, vision alone is often not sufficient to properly execute dexterous manipulation tasks.
A representation learned using multiple sensor modalities, e.g. vision, touch and proprioception, can be fundamental to capture key features of objects.

\begin{figure}
\centering
\includegraphics[width=\textwidth]{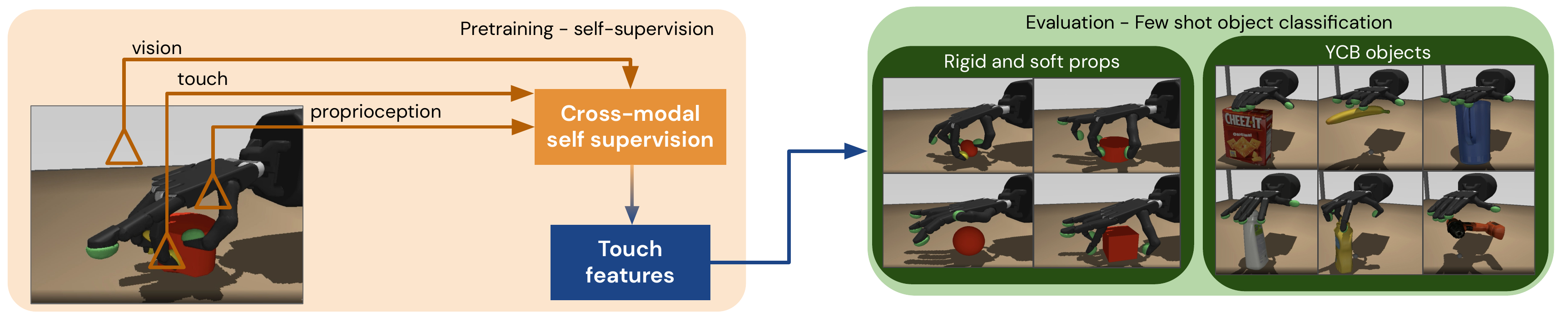}
\caption{Learning touch features though cross-modal self-supervision and evaluation on few-shot object classification.
Learned features can in turn 
be used to identify different objects and their properties across diverse sets of synthetic and real-world objects.\vspace*{-5mm}}
\label{fig:cm_touch_overview}
\end{figure}

In this paper, we present an extensive study of 
cross-modal 
techniques that learn rich representations from  vision and touch, without requiring external supervision.
We apply many existing multi-modal representation learning methods, such as $L^3$-net~\citep{arandjelovic2017look}, CMDC~\citep{aytar2018playing}, TCN~\citep{sermanet2018time} and CMC~\citep{tian2019contrastive}, to vision and touch modalities. We develop two cross-modal versions of CPC~\citep{oord2018representation} and implement a cross-modal generation method (CM-GEN), which generates the visual observations using the touch features. We evaluate the effectiveness of learned touch representations on few-shot classification of objects and their orientations with increasing levels of complexity. A visual representation of our setup is shown in Fig.~\ref{fig:cm_touch_overview}. We demonstrate that learned touch features are significantly better than using raw proprioception and touch features, particularly in few-shot learning scenarios where we obtain a 25\% average performance boost.
We release two datasets\footnote{accessible at \url{https://github.com/deepmind/deepmind-research/tree/master/cmtouch}}
with the rich multimodal data we trained on, to enable further research in cross-modal representation learning with touch and vision. These are based on a simulated Shadow Dexterous hand~\citep{ShadowHand}, and include visual, tactile and proprioceptive trajectories.
\vspace*{-.2cm}

\section{Related work}
\label{sec:relatedwork}
\vspace*{-.2cm}
When it comes to robotics, the importance of tactile sensing has long been acknowledged \citep{yousef2011tactile, dahiya2009tactile, hammock201325th, cramphorn2016tactile} but making use of it has proven challenging for a number of reasons, from hardware related issues like wear and tear and drift, to the lack of simple ways to specify tactile-based goals or trajectories, as well as to the limitations of physics models of contacts and forces.
Recent works have explored methods to combine vision and touch to learn representations from robot interactions with objects~\citep{lee2019touching, yuan2017connecting, zheng2020lifelong, lin2019learning, zambelli2020multimodal} and several methods have been proposed to learn multi- and cross-modal representations from large datasets~\citep{aytar2017cross, ngiam2011multimodal, lin2020interbert, lu2019vilbert, le2019improving, owens2016visually, owens2016ambient, aytar2016soundnet} in other machine learning domains, but a comprehensive evaluation of how complete and useful these representations are on practical tasks is still lacking.

\vspace*{-.2cm}
\paragraph{Cross-modal learning} 
Our work is mostly related to cross-modal representation learning~\citep{aytar2017cross, ngiam2011multimodal}, which aims to generate a representation that effectively correlates different sensor modalities.
Several studies explored cross-modal transfer between images and text~\citep{socher2013zero}, as well as models for vision and language representations~\citep{lin2020interbert, lu2019vilbert}.
Interesting work has been done on audio-visual representations \citep{le2019improving, owens2016visually, owens2016ambient, aytar2016soundnet} 
and on generating captions for images \citep{donahue2015long, karpathy2015deep, kulkarni2013babytalk, xu2015show}.
Large-scale cross-domain datasets were fundamental to their success, but these types of datasets are currently unavailable for vision and touch. We hence used a robotic platform equipped with tactile sensors to collect data for our study.

We are specifically interested in the interplay between vision and touch.
Recent studies have addressed the problem of integrating these sensor modalities 
by studying the 
physical properties of fabrics via visual and tactile data fusion~\citep{yuan2017connecting, zheng2020lifelong, lee2019touching}, or by focusing  
on multimodal learning using proprioception, vision, touch and sound~\citep{liu2018cross,zambelli2020multimodal,zambelli2016online, calandra2017feeling}.
Different from prior work that 
addressed specific manipulation tasks, we focus on several cross-modal representation methods to learn touch features through self-supervision.
Rather than shaping the learned representation around a specific downstream manipulation task (such as picking-up~\citep{chan2019effects}, grasping~\citep{calandra2017feeling}, classification~\citep{falco2019transfer,lee2019touching}), the proposed approach focuses on learning a representation independent of the target use case using self-supervision. \vspace*{-.1cm}

\section{Learning Rich Touch Features}
\label{sec:learning_touch}
\begin{figure}[pt]
\centering
\includegraphics[width=.9\textwidth]{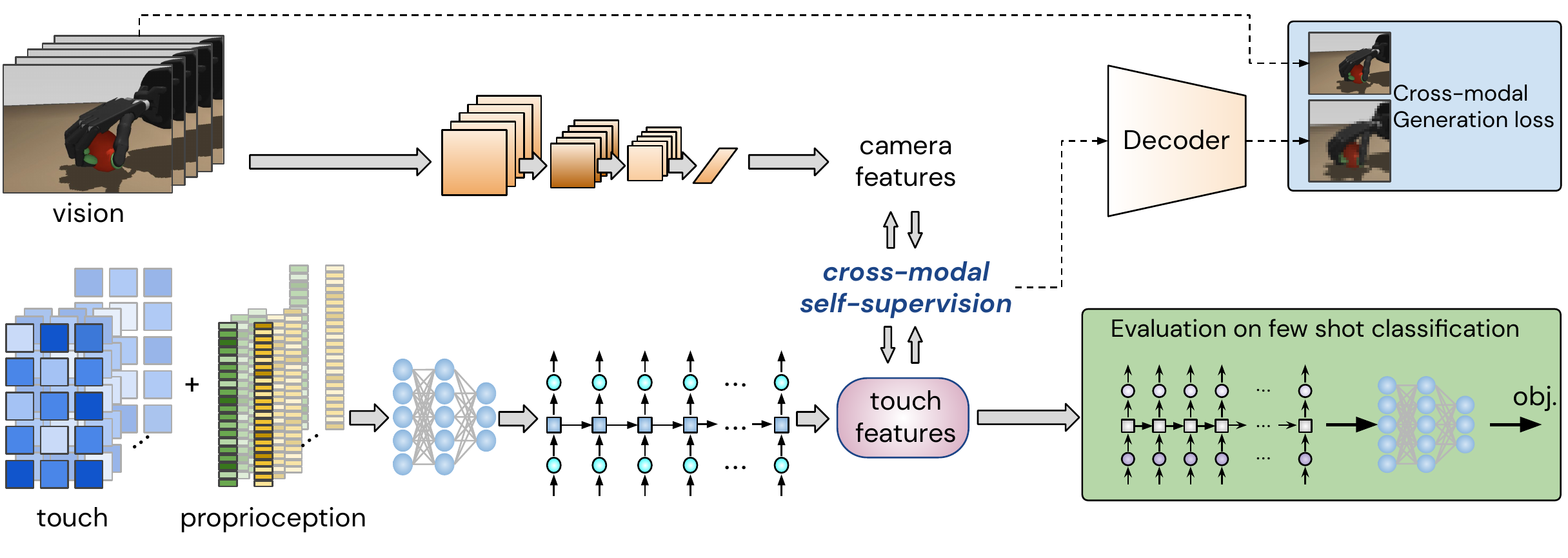}
\caption{Overview: self-supervised cross-modal representation learning of touch features. Cross-modal generation loss used to train features  (top-right), and evaluation networks (bottom-right). \vspace*{-.2cm}}
\label{fig:models}
\end{figure}

\vspace*{-.2cm}
In this section we describe how high-level touch features are learned by exploiting naturally existing correlations between touch perception and vision. 
No annotations or supervision is provided for feature learning, i.e., the process is entirely \emph{self-supervised}. As described in Section~\ref{sec:methodology}, our episodic datasets of robotic hand and object interactions are obtained by running a policy trained blindly to explore its surroundings. The features are learned by capitalizing on temporal correlations of vision and touch within the provided episodes (Fig.~\ref{fig:models}).

Formally, for any given paired vision ${\mathbf{o}=\{o_i\}_{i=1}^{N}}$ and touch ${\mathbf{t}=\{t_i\}_{i=1}^{N}}$ sequences, where $o_i$ and $t_i$ are visual and touch observations at time step $i$, we first encode them to obtain the vision and touch features  $v_i = f(o_i)$ and $x_i = g(t_i)$ respectively, and then we employ several cross-modal self-supervision objectives to learn latent touch and vision features.
Next we'll present the touch and vision encoders, then we'll describe the cross-modal self-supervision objectives used for learning touch features.\vspace*{-3mm}

\subsection{Encoders}\vspace*{-.2cm}
\noindent{\bf Touch Encoding.} Touch and proprioception are encoded through a 4-layers MLP followed by a LSTM network, which allows to accumulate the dynamics information provided by tactile sensors and robot joints throughout an episode. Capturing this information in a time window is critical to learn meaningful features that reflect the dynamical nature of these signals.

\noindent{\bf Vision Encoding.} Visual input is obtained from simulated cameras, and is processed through a convolutional encoder with residual connections. Each frame is processed separately. Contrary to proprioception and touch information, visual information is not accumulated with any memory unit.

\noindent{\bf Implementation details.} 
All implementation details are reported in the Supplementary material.
\vspace*{-.2cm}

\subsection{Cross-modal self-supervision}\vspace*{-.2cm}
We adapted several cross-modal self-supervision objectives, such as $L^3$-net\citep{arandjelovic2017look}, CMDC\citep{aytar2018playing}, TCN\citep{sermanet2018time}, and CMC\citep{tian2019contrastive} to learn touch and vision features. We developed cross-modal versions of contrastive predictive coding (CPC)\citep{oord2018representation} with two variants. We also implemented a cross-modal generation method CM-GEN which generates the visual observations using the touch features. All the methods operate over the latent paired sequences $\mathbf{v}=\{v_i\}_{i=1}^{N}$ and $\mathbf{x}=\{x_i\}_{i=1}^{N}$ except for the CM-GEN which also uses $o_i$ while generating visual frames. We describe each of these methods below 
using a single paired vision and touch sequence, but we optimize them with all the episodes in our dataset.

\noindent{\bf $\mathbf{L^3}$-net\citep{arandjelovic2017look}}. The main idea in $L^3$-net is binary classification of temporally sync and non-sync pairs in multiple modalities. In our case a positive example is a synced pair $(v_i,\ x_i)$ and negative examples are any non-sync pairs $(v_i,\ x_j),\ \forall i \neq j$. Non-sync pairs are also acquired by mixing vision and touch latents from multiple episodes. Given a positive or a negative pair, we concatenate touch and vision features $(v,\ x)$ and run them through a two-level MLP to perform binary sync/non-sync classification. The loss function is binary cross-entropy.

\noindent{\bf CMDC\citep{aytar2018playing}}. Instead of directly using synced pairs, cross-modal distance classification (CMDC) predicts the temporal distance between the given cross-modal pair. For instance in the pair $(v_i,\ x_{i+1})$ the temporal distance is $1$. CMDC sets distance prediction as a classification problem where the classes are different distance intervals. We follow the same setting described in \citep{aytar2018playing}. Given $(v,\ x)$ we run them through a two-layers MLP to perform distance classification  with a softmax cross-entropy loss.

\noindent{\bf TCN\citep{sermanet2018time} and CMC\citep{tian2019contrastive}}. Time contrastive networks (TCN) learn representations by enforcing higher similarity between sync pairs across two different camera views compared to any pair selected within a single view sequence. In our context two views are the vision and touch modalities. We particularly adopted n-pairs implementation \citep{sermanet2018time} as opposed to the triplet implementation as the model learns to cheat in triplet implementation when there is an LSTM in one of the modalities (i.e. touch). Given the paired sequences $\mathbf{v}$ and $\mathbf{x}$ the TCN $n$-pairs objective that we minimize is:
\begin{equation}
 -\log \, \frac{\exp(x_i^\intercal v_i)}{\sum_{j}\exp(x_i^\intercal v_j)} \label{tcn_npairs}
\end{equation}
where $(x_i, v_i)$ is the positive pair and  $(x_i, v_j), \forall j \neq i$ are the negative pairs. This specific implementation has very strong connection to contrastive multi-view coding (CMC\citep{tian2019contrastive}) and multi-modal noise-contrastive estimation (NCE\citep{alayrac2020self}) which are essentially the same objectives. The aim in both methods is to classify the matching pair among all other non-matching negative pairs. The main difference compared to TCN is that negative pairs are also populated from other sequences in the batch. In  
equation (\ref{tcn_npairs}) we can also swap vision and touch to obtain the symmetric loss function. In our implementations of TCN and CMC we use the symmetric losses as well.

\noindent{\bf CPC\citep{oord2018representation}}. Contrastive predictive coding (CPC) predicts the future latent observations using the current latent observation and previous context which is accumulated through an LSTM. It is originally developed for single modal sequences. Here we build a cross-modal version where a touch feature $x_i$ predicts $n$-step into the future in the vision modality. Each of the $n$-step predictions are obtained through $n$ linear predictors for each time step distance. The objective is formalized as a contrastive learning approach where $(v_i, \hat{v}_i)$ is the positive pair and $\hat{v}_i$ is the predicted vision feature using an earlier touch feature. Note that we have $n$ different predictions of $v_i$ as any previous $x_{i-1}, x_{i-2}, ..., x_{i-n}$ can provide their own prediction $\hat{v}_i$. The minimized objective is:
\begin{equation}
 -\log \, \frac{\exp(\hat{v}_i^\intercal v_i)}{\sum_{j}\exp(\hat{v}_i^\intercal v_j)} \label{cpc}
\end{equation}
where we contrast the matching prediction with all other non-matching predictions. We refer to this method as cross-modal CPC (CM-CPC) as it is a straightforward adaptation of CPC to multiple modalities. Unlike the regular CPC, as we operate in multiple modalities we can also predict the current or previous latents in the other modality. Hence we also introduce 
CM-CPC-H, a variation of CM-CPC that predicts the $n$-step history, current, and $n$-step future latents in the vision modality. In practice we set $n=20$, similar to \citep{oord2018representation}. For more details on CPC please refer to \citep{oord2018representation}.

\noindent{\bf CM-GEN}. Generating one modality from another is performed numerous times in multiple domains for different purposes \citep{isola2017image, chen2017deep, chen2017photographic, salvador2019inverse}. In our context we generate vision from touch mainly for learning high-level touch representations. We employ a 5 layers convolutional decoder $d$ to the touch feature $x_i$ to generate the corresponding visual frame $o_i$. The minimized loss is  $||d(x_i) - o_i||^2$ for all the time steps $i$. The details of the decoder can be found in the Supplementary material.

\noindent{\bf Implementation details.} 
We use Adam~\citep{kingma2017adam}, with learning rate $0.001$ for CM-GEN and $0.0001$ for all other methods.
All the other implementation details are reported in the Supplementary material.
\vspace*{-.2cm}

\section{Dataset and Experimental Setup}
\label{sec:methodology}
\vspace*{-.2cm}
The proposed approach allows us to learn touch features through cross-modal self-supervision (Fig.~\ref{fig:models}). The learned features are then evaluated on few-shot classification tasks, to identify different objects and their properties.
The following paragraphs describe how the datasets have been generated.

\vspace*{-.1cm}
\paragraph{Experimental setup}
We run experiments in simulation with MuJoCo~\citep{Todorov_2012} 
and we use the simulated Shadow Dexterous Hand~\citep{ShadowHand}, with five fingers and 24 degrees of freedom, actuated by 20 motors. 
In simulation, each fingertip has a spatial touch sensor attached with a spatial resolution of $4\times4$ and three channels: one for normal force and two for tangential forces. We simplify this by taking the absolute value and then summing across the spatial dimensions, to obtain a 3D force vector for each fingertip. 
More details about the tactile sensors are reported in the Supplementary material. 

The state consists of proprioception (joint positions and joint velocities) and touch. 
Visual inputs are collected with a $64\times64$ resolution and are only used for representation learning, but are not provided as observations to control the robot's actions.
The action space is 20-dimensional. We use velocity control and a control rate of 30 Hz. 
Each episode has 200 time steps, which correspond to about 6 seconds. 

The environment consists of the Shadow Hand, facing down, and interacting with different objects. These objects have different shapes, sizes and physical properties (e.g. rigid or soft). 
We develop two versions of the task, the first using simple props and the second using YCB objects. In both cases, objects are fixed to their frame of reference, while their position and orientation are randomized.
Focusing on this simple environment enables us to clearly characterize the effectiveness of our methods for training cross-modal and touch-based representations. 

\begin{figure}[]
\centering
\includegraphics[width=\textwidth]{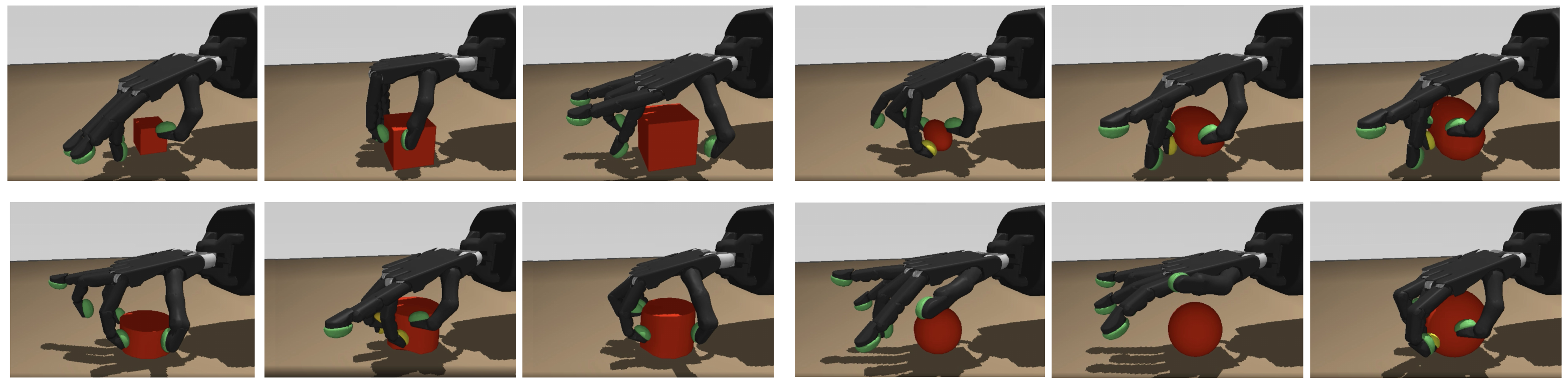}
\caption{Samples from the Props dataset: the Shadow hand is presented with objects sampled from four shapes of different sizes, and they can be either rigid or deformable objects.\vspace*{-4mm}}
\label{fig:sample_props}
\end{figure}

\vspace*{-.1cm}
\paragraph{Props}
Props are simple 3D shaped objects that include cubes, spheres, cylinders and ellipsoid of different sizes.
We also generated the soft version of each prop, which can deform under the pressure of the touching fingers.
Soft deformable objects are complex entities to simulate: they are defined through a composition of multiple bodies (capsules) that are tied together to form a shape, such as a cube or a sphere. The main characteristic of these objects is their elastic behaviour, that is they change shape when touched. The most difficult thing to simulate in this context is contacts, which grow exponentially with the increased number of colliding bodies.
Forty-eight different objects are generated by sampling from 6 different sized, 4 different shapes (i.e. sphere, cylinder, cube, ellipsoid), and they can either be rigid or soft.
A sample from this dataset is shown in Fig.~\ref{fig:sample_props}.

\begin{figure}[]
\centering
\includegraphics[width=\textwidth]{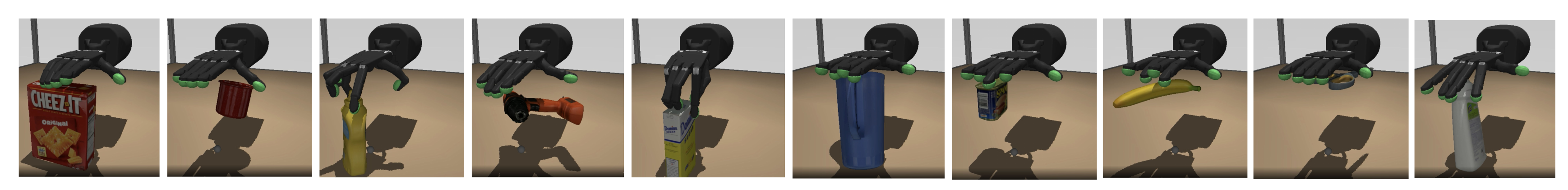}
\caption{Samples from the YCB objects dataset: the Shadow hand is presented with objects sampled from ten classes, which appear in different orientations and poses.\vspace*{-4mm}}
\label{fig:sample_ycb}
\end{figure}

\vspace*{-.1cm}
\paragraph{YCB objects}
The YCB objects dataset \citep{calli2017yale} consists of everyday objects with different shapes, sizes, textures, weight and rigidity.
We chose a set of ten objects:
cracker box, sugar box, mustard bottle, potted meat can, banana, pitcher base, bleach cleanser, mug, power drill, scissors.
These are generated in simulation at their standard size, which is also proportionate to the default dimension of the simulated Shadow Hand (see examples in Fig.~\ref{fig:sample_ycb}).
The pose of each object is randomly selected among a set of 60 different poses, where we vary the orientation of the object. These variations make the identification of each object more complex and require a higher generalization capability from the learning method applied. More details about the dataset generation are reported in the Supplementary material.
A sample from this dataset is shown in Fig.~\ref{fig:sample_ycb}.

\paragraph{Control}
A Maximum a Posteriori Policy Optimization (MPO) \citep{abdolmaleki2018maximum} reinforcement learning (RL) agent is used to control the hand movements. At the beginning of each episode, an object is chosen randomly among the set of objects. The agent is rewarded when touching the object with the sensorized fingertips: the more fingertips are in contact with the object, the higher the reward. The agent is blind, as no visual input is fed to its networks, and controls the hand by issuing velocity commands.
It is important that no visual information is provided to the policy: when vision is present as input to the agent, visual clues are exploited to generate control trajectories. In turn, the data collected, specifically proprioception, would contain significant biases leaked from the visual clues given to the policy. By training a blind agent we make sure that proprioception and touch are not biased by any visual clue.\vspace*{-3mm}

\section{Experimental Results}
\label{sec:results}
\vspace*{-.1cm}
In this section we include quantitative results of few-shot classification, evaluation on unseen objects and pose estimation. A discussion follows.\vspace{-2mm}

\begin{table}[]\centering
\caption{Few-shot classification - Props and YCB dataset. 
}
\label{tab:few_shot}
\includegraphics[width=\linewidth]{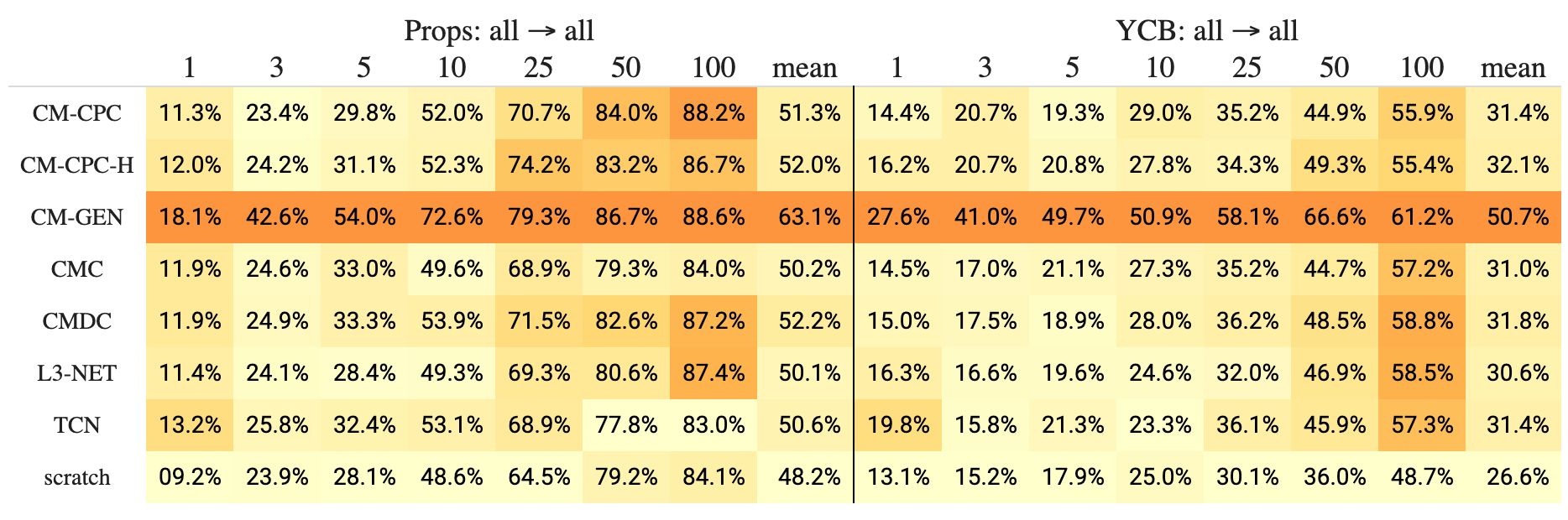}\vspace*{-.5cm}
\end{table}

\vspace*{-.2cm}
\paragraph{Few-shot classification}
We evaluate all our training methods on few-shot classification from the learned touch features. Vision is not available at this time, just as when looking for the keys in a bag.
We compare our self-supervised approach with a learning from scratch approach, where features are learned directly from touch and proprioception while optimizing for the classification objective. 
For the Props dataset, a class is defined by an instance of a specific shape (cube, sphere, cylinder or ellipsoid), size (one of 6 different sizes) and rigidity (rigid or soft object). There is therefore a total of 48 classes for the props dataset. There are instead 10 classes for the YCB objects dataset, corresponding to the 10 different objects used to collect the data. Classification of the 10 YCB objects is nonetheless challenging due to the variations applied to the pose of each object in each episode.

Results of few-shot classification are presented in Table~\ref{tab:few_shot}. Learned features are learned on the entire datasets, and are evaluated on few-shot classification using 1, 3, 5, 10, 25, 50 or 100 examples for each object class. The mean across scores is also reported. Darker colors indicate better scores. 
 The standard deviation obtained across multiple runs of the experiment is between 0.01 and 0.04.

\begin{figure}[]\centering
\begin{minipage}{.455\textwidth}
\includegraphics[trim=0 0 440 0, width=\linewidth]{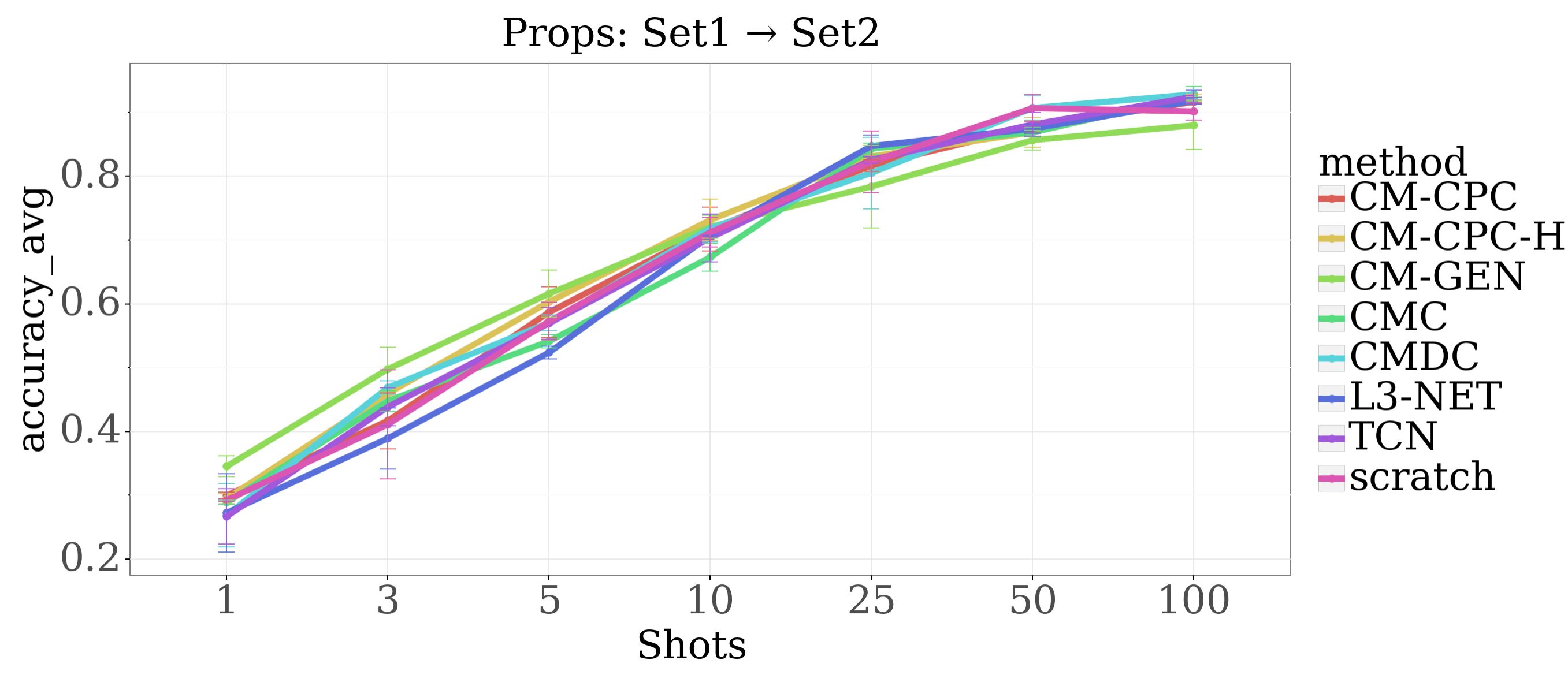}
\end{minipage}%
\begin{minipage}{.545\textwidth}
\includegraphics[width=\linewidth]{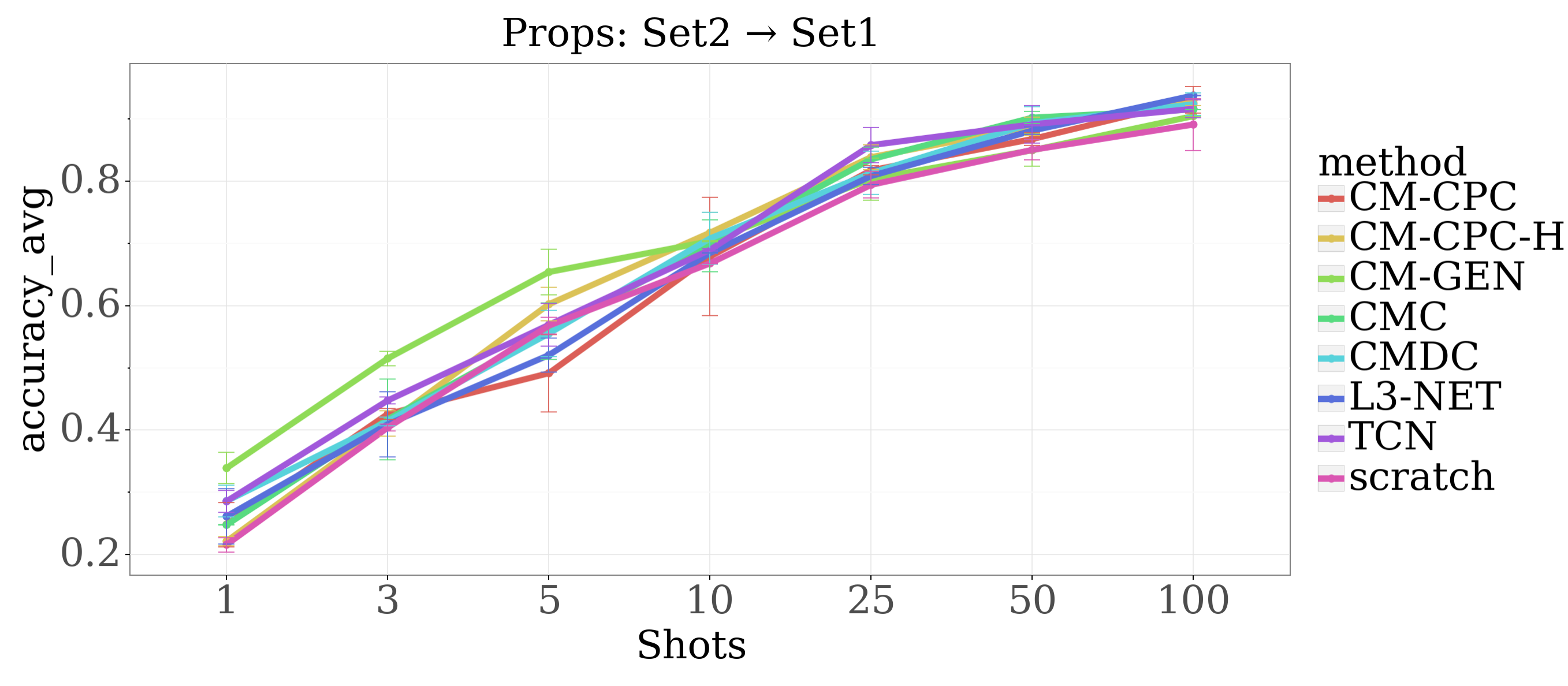}
\end{minipage}
\vspace*{-3mm}
\caption{Evaluation on unseen objects - Props objects classification. In the low shots regime CM-GEN outperforms other methods, but above $10$ shots the difference is less prominent, with CMDC and TCN showing slightly better performance than other methods in average.}
\label{fig:props_transfer}
\end{figure}

\begin{figure}[]\centering
\begin{minipage}{.455\textwidth}
\includegraphics[trim=0 0 440 0, width=\linewidth]{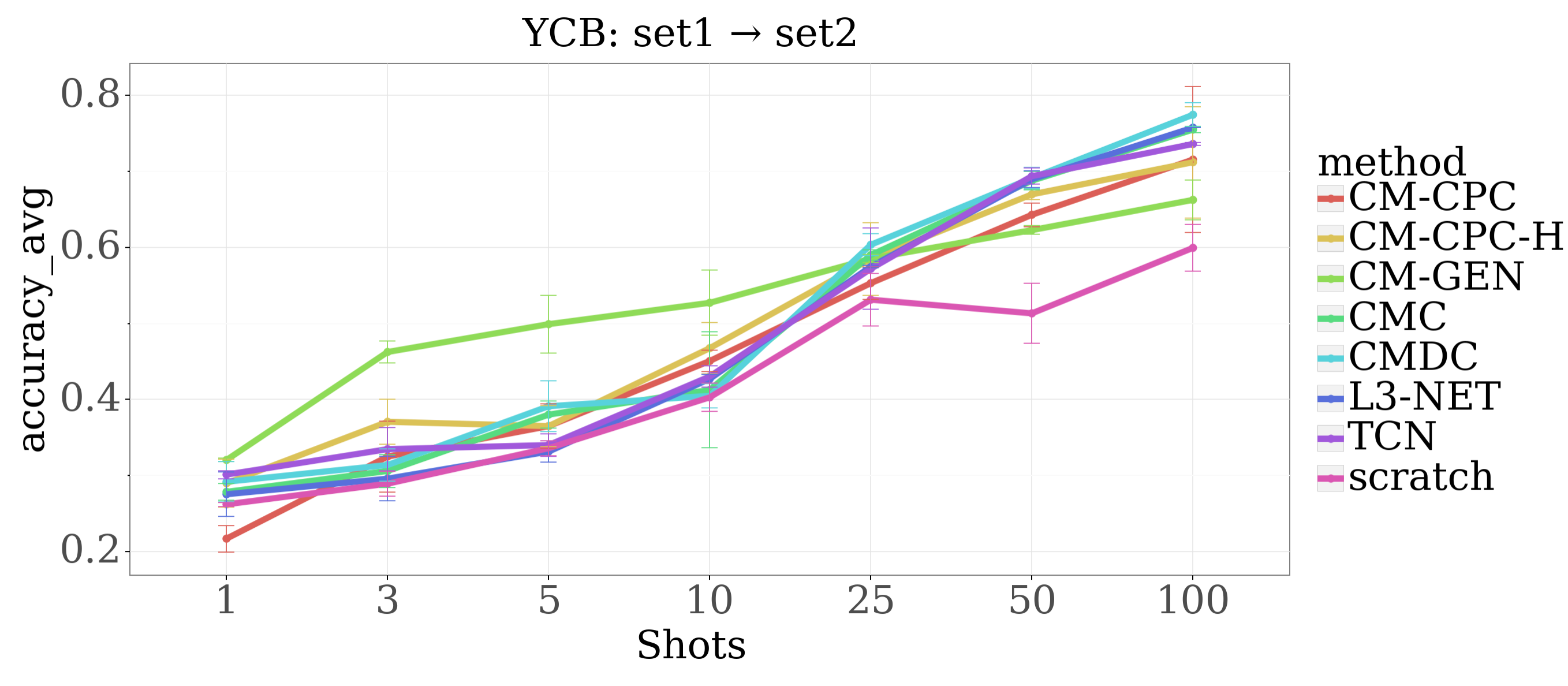}
\end{minipage}%
\begin{minipage}{.545\textwidth}
\includegraphics[width=\linewidth]{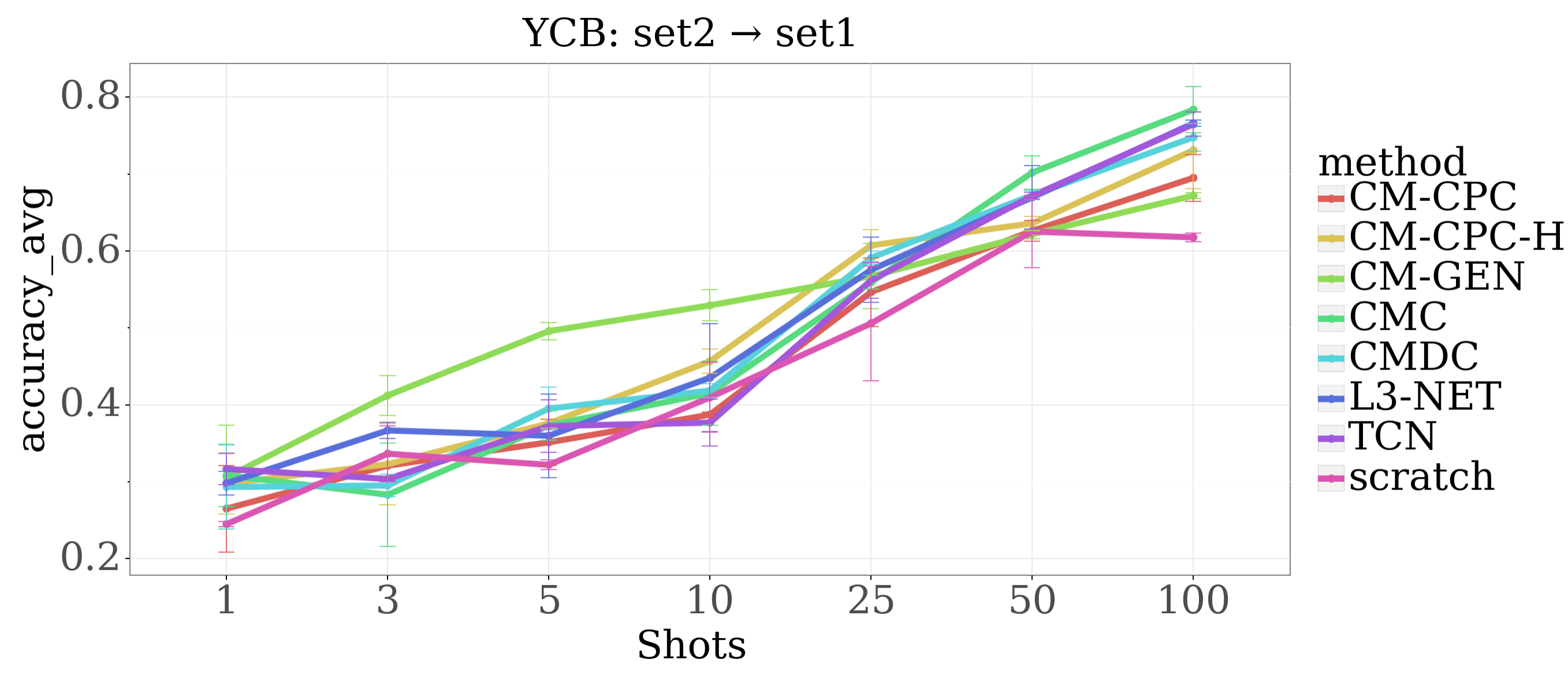}
\end{minipage}
\vspace*{-3mm}
\caption{Evaluation on unseen objects - YCB objects classification. Similar to the results on the Props dataset, in the low shots regime CM-GEN outperforms other methods. Above the $25$ shots regime, CMC, CMDC, TCN and L3-NET tend to perform best with similar performance.}
\label{fig:ycb_transfer}
\end{figure}

\vspace*{-.2cm}
\paragraph{Evaluation on unseen objects}
We evaluate learned touch features on objects that were not presented in the pretraining step. In order to do so, we generate two disjoint sets of objects from the collected datasets, that we call simply Set1 and Set2. One of the sets is used to train the self-supervised representation, and the other set is used to evaluate the learned touch features. The two sets are randomly generated to be disjoint, that is Set1 and Set2 don't have any object in common, hence models are evaluated on completely unseen objects.
Results of evaluation on unseen objects from Set1 to Set2 and from Set2 to Set1 are presented in Figures~\ref{fig:props_transfer} and~\ref{fig:ycb_transfer} for the Props and YCB objects datasets, respectively. 
 Similar to the previous experiment, the standard deviation obtained across multiple runs of this experiment is between 0.01 and 0.04. 
Further results are reported in the Supplementary material, including all accuracy scores of the different methods.

\begin{figure}[]\centering
\begin{minipage}{.5\textwidth}
\includegraphics[trim=0 0 420 0, width=\linewidth]{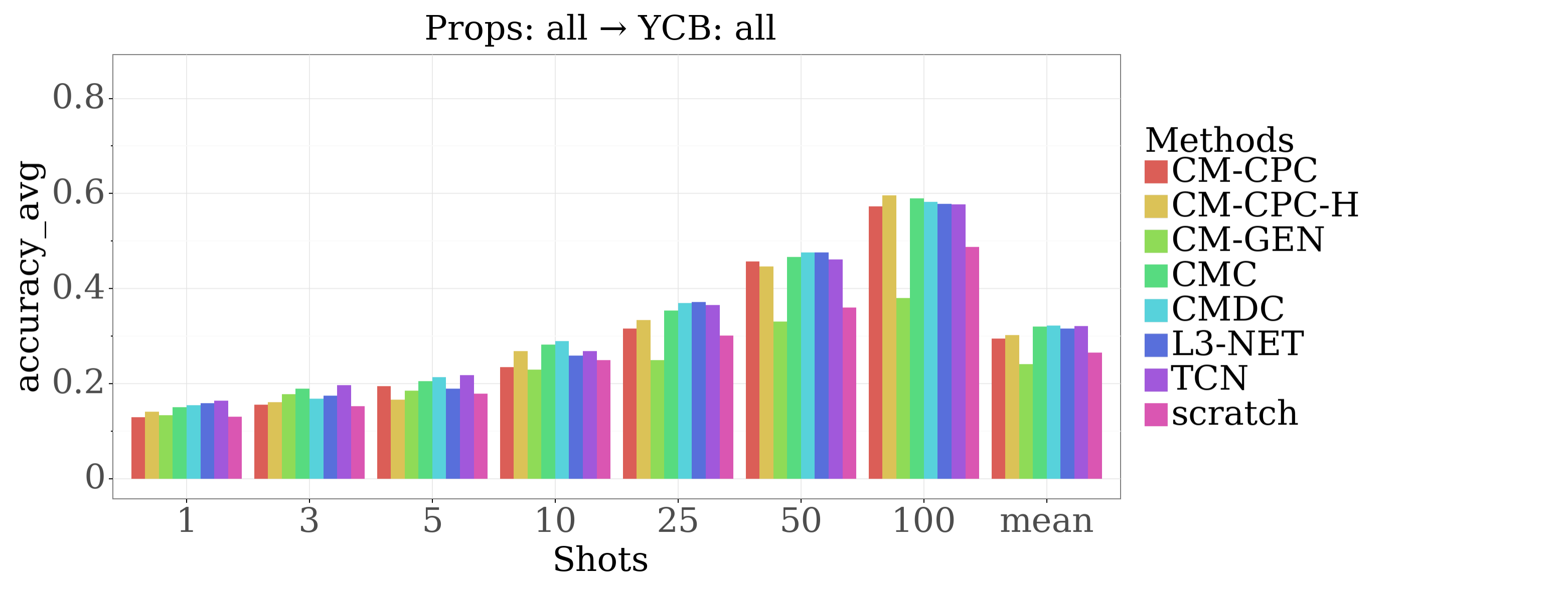}
\end{minipage}%
\begin{minipage}{.5\textwidth}
\includegraphics[width=\linewidth]{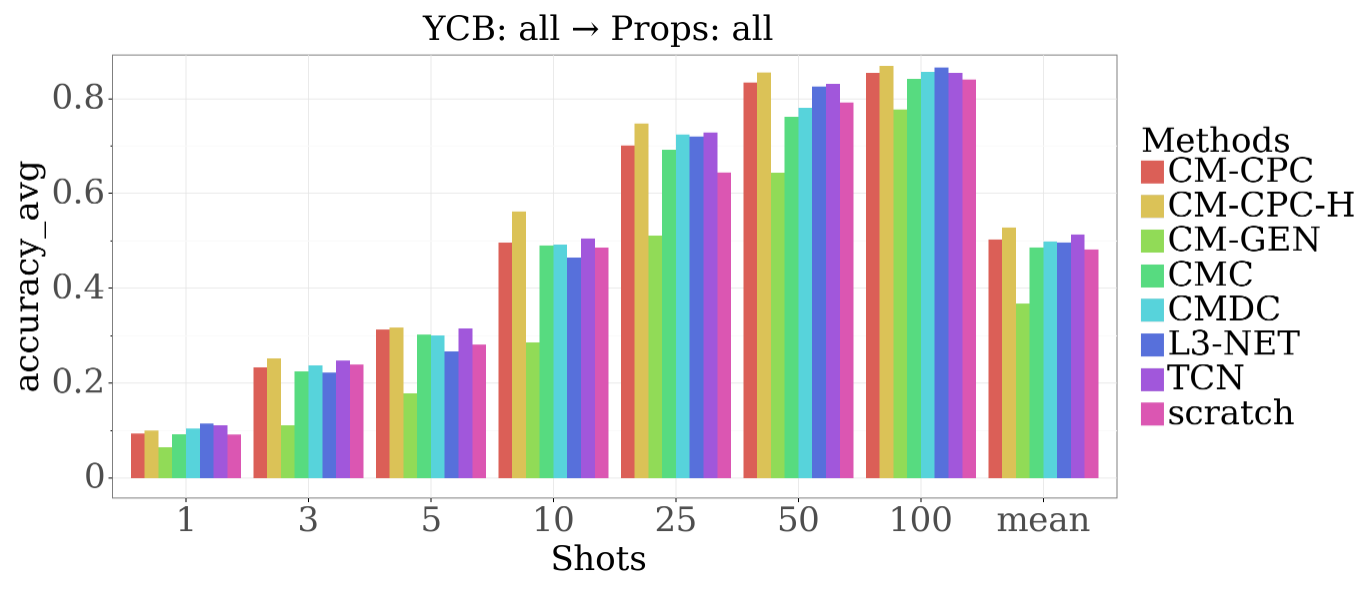}
\end{minipage}
\vspace*{-3mm}
\caption{Evaluation on unseen objects across datasets.}
\label{fig:transfer}
\end{figure}

We also perform a cross-dataset evaluation: touch features learned on the props datasets are evaluated on few-shot classification of YCB objects, and viceversa. Results of this evaluation are presented in Fig.~\ref{fig:transfer}.
For both sets of experiments, learned features are trained on one of the datasets, and evaluated on few-shot classification on the other dataset, using 1, 3, 5, 10, 25, 50 or 100 examples for each object's class. 

\vspace*{-.2cm}
\paragraph{Pose estimation}
For the YCB objects dataset, we also evaluated the learned touch features on estimating the orientation of each object class. Pose estimation is formulated as a classification across a set of different poses that an object can take:
each object can appear in 60 different poses, given by a rotation around the vertical axis and a tilt from this same axis.

\begin{figure}[!t]
\begin{minipage}{\textwidth}
\begin{minipage}[b]{0.45\textwidth}
\centering
\includegraphics[width=\linewidth]{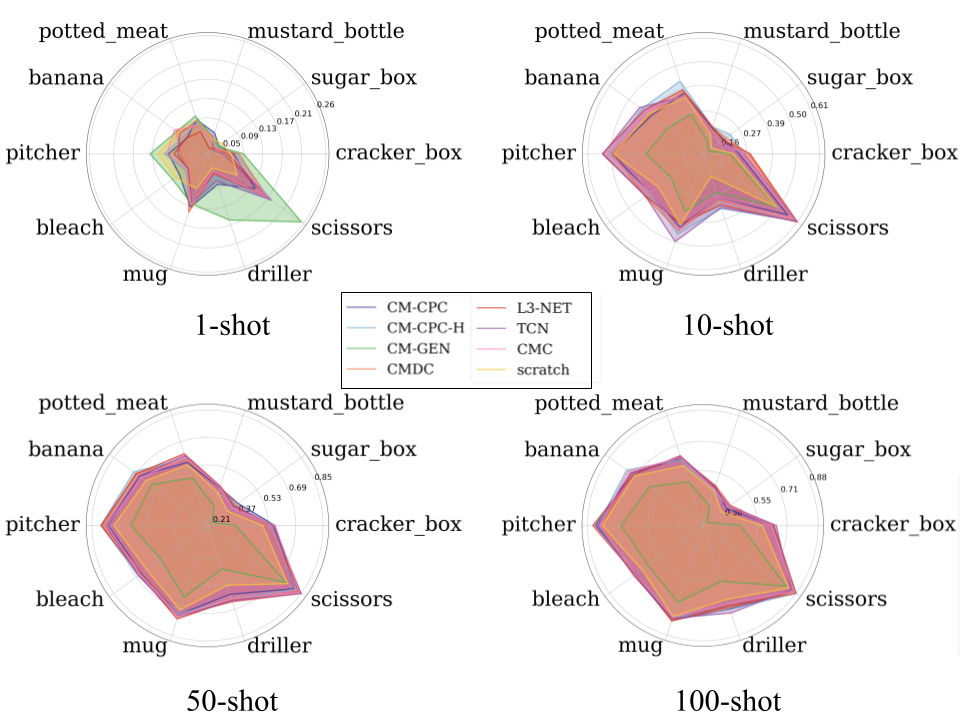}
\captionof{figure}{Pose estimation accuracy of YCB objects: 1-shot, 10-shot, 50-shot and 100-shot classification. Larger area is better.}
\label{fig:pose_few_shot}
\end{minipage}
\hfill
\begin{minipage}[b]{0.49\textwidth}
\centering
\includegraphics[trim=0 0 0 68, clip, width=\linewidth]{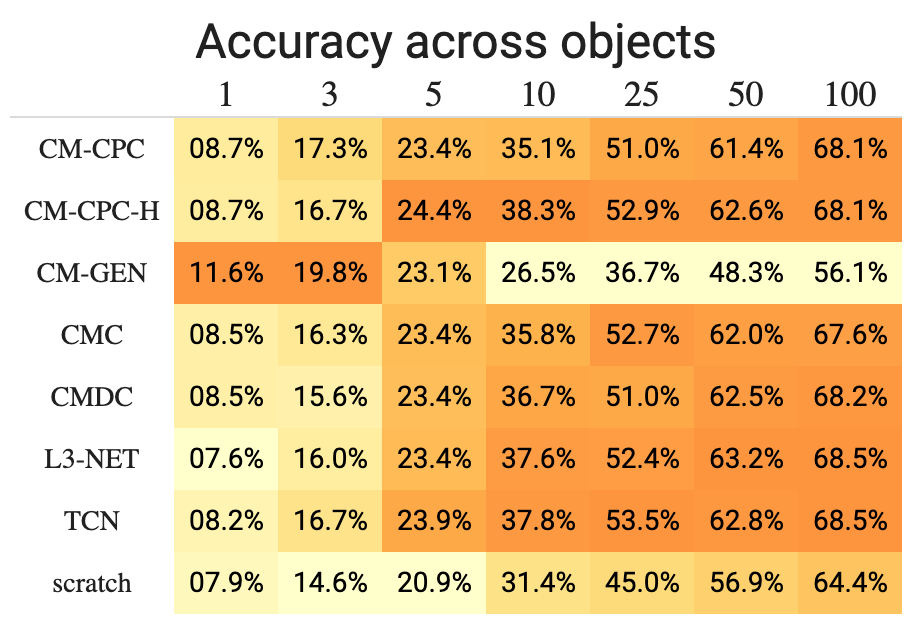}
\captionof{table}{Pose estimation: accuracy across all objects. 
Learned features are evaluated on few-shot classification using 1, 3, 5, 10, 25, 50 or 100 examples for each object's class. Darker colors indicate better scores.}
\label{tab:pose_acc_across_objs}
\end{minipage}
\end{minipage}
\vspace*{-.5cm}
\end{figure}

Results of few-shot pose estimation are presented in Fig.~\ref{fig:pose_few_shot}, where for each object we report the accuracy obtained with only 1, 10, 50, 100 examples of that object. 
 Table \ref{tab:pose_acc_across_objs} reports the average accuracy across all objects, for all methods and all few-shots experiments. 
Further results are reported in the Supplementary material, including all few-shot (1, 3, 5, 10, 25, 50 or 100) experiments.

\vspace*{-.2cm}
\subsection{Discussion}
\vspace*{-.2cm}
Results presented above show that cross-modal self-supervision improves touch features: touch features learned from scratch achieved lower scores compared to features learned through our proposed cross-modal self-supervision approach.

In all experiments, as expected, we notice an increase in performance when more examples are provided for $n$-shot classification (i.e., when $n$ is higher). The accuracy is low when only a single example is provided ($18.1\%$ for props and $27.6\%$ for YCB objects - Table~\ref{tab:few_shot}), and grows with the number of seen examples, reaching an accuracy of $88.6\%$ and $61.2\%$ for props and YCB objects, respectively, when 100 examples are available. Note that props are generally easier to identify compared to YCB objects, given the simple shapes and the different textures (rigid or soft).

For few-shot classification evaluated on single dataset, CM-GEN performed best compared to other self-supervision methods. This is because visual clues captured during self-supervision are sufficient and effective. 
Other methods become competitive when training is performed on a set of objects and the learned features are then evaluated on a second set. In this case, results confirm that cross-modal self-supervision achieves considerably better performance than training touch features from scratch. CM-GEN still achieves slightly better results in the low data regime (when only a small number of classes are observed), while other methods based on contrastive approaches (e.g. TCN and $L^3$-NET) achieve better results with 50 or 100 examples (Figures~\ref{fig:props_transfer} and ~\ref{fig:ycb_transfer}).
We also note that CM-GEN overfits to the dataset it is trained on, achieving the worst performance in the identification of objects that belong to a different dataset (Fig.~\ref{fig:transfer}). In this case, transferring from one group of objects to the other requires a higher level of abstraction, which does not necessarily favour visual clues.
These results indicate that methods based on contrastive self-supervised losses, such as TCN, $L^3$-NET, CMDC and CMC can learn more robust touch features compared to CM-GEN and thus achieve the best overall performance.

Finally, the proposed approach achieves good performance also on the pose estimation experiment, where the classification is harder due to the large number of classes per object. As observed above, cross-modal generation performs poorly on this task, due to the increased complexity: discriminating orientation from vision is hard and doesn't provide features that are robust enough. On the other hand, all other self-supervised methods achieve better performance compared to learning features from scratch, thus showing that the proposed approach is an effective way to improve touch representation for manipulation skills with different objects.\vspace*{-3mm}

\paragraph{Validation on real data} While no real world data have been used in these experiments, the proposed approach is agnostic to the specific kind of data used: no bias is introduced in the data format, and the encoder networks used to train the self-supervised representation can be easily adapted, if needed, to accommodate inputs from a real robotic platform. More details in the Supplementary material.
\vspace*{-.3cm}


\section{Conclusion}
\label{sec:conclusion}
\vspace*{-.25cm}

The use of touch is bound to have a critical role in robotics manipulation, especially in tasks involving dexterity, manipulation of daily-life objects, dealing with occlusions or closed and constrained environment. 
We took inspiration from a challenging task: finding an object in a bag. In this paper we present an approach to use touch and vision to build representations that allow to identify objects from touch and proprioception alone.
We benchmark several self-supervised learning methods on three challenging tasks: few-shot classification, knowledge transfer to unseen objects, and pose estimation from touch.
Experimental results show that cross-modal self-supervision enables the learning of rich touch features that are effective on the final task. In fact, our evaluations demonstrate that touch features learned through cross-modal self-supervised approaches improve classification performance on few-shot and unseen objects classification, as well as pose estimation of several real-world objects.

Furthermore, to spark interest in these types of tasks and to foster further research on the use of touch in robotic manipulation, we release two rich multimodal datasets used in this work.

\small
\bibliography{references}
\normalsize
\linespread{1.}



\section*{Supplementary material}

# *Supplementary material*

## 1  Considerations on CoRL2020 and the Pandemic

During the few months before the CoRL2020 deadline, we did not have a reach to our real robots, due to the lockdown that followed the pandemic outbreak. This implied that we could not carry on our experiments (in particular the data collection) on the real robotic platform. We were still able to collect that data in simulation, using Mujoco, a simulated Shadow hand, and a set of diverse objects, both synthetic (generated by us) and externally available realistic objects (i.e. the YCB objects).

Despite the limitations on the data collection, the main contributions of our paper still hold: the proposed approach is mainly related to perception, and it is agnostic of the specific data source that is used for training. All the networks and architecture used for self-supervision can be easily adapted, if needed, to real world data. Our approach can be applied to data coming from a real robot setup, provided that tactile data (of some form) can be collected, to learn touch features.

## 2  Dataset generation

**Props**  Props are generated in Mujoco as entities of four different shapes: cube, sphere, cylinder, ellipsoid. Each shape is generated by setting either a height and/or a radius, sampling from $\{0.030, 0.035, 0.040, 0.045, 0.050, 0.055\}$cm.

To generate deformable versions of these props we construct composite entities starting from simple shapes: each soft prop is composed of several elements or sub-body, called capsules. Each capsule is built by defining its dimensions (length and radius), and a soft prop is then characterized by the number of capsules that will form the prop itself: the more capsules, the denser the prop; the bigger the capsules, the bigger the prop.

As explained in the paper, we generated two sets of data including disjoint sets of props.  We chose random disjoint subsets and fixed the selections for repeatability. The two sets are composed as follows, where each number correspond to an object's class: $\text{Set}1 = [29, 4, 26, 30, 32, 37, 34, 40, 7, 10, 11, 31, 33, 27, 47, 2, 46, 18, 15, 28, 22, 16, 41, 20]$, $\text{Set}2 = [42, 8, 13, 25, 5, 17, 35, 14, 38, 1, 12, 43, 24, 6, 23, 36, 21, 19, 9, 39, 45, 3, 0, 44]$. Each object's class is computed from a triplet of codes identifying the shape, the size and the rigidity (binary) of each prop. There are on average 1700 instances for each class.

**YCB objects**  YCB objects are imported as entities in Mujoco from the open source dataset. Each object is imported with its standard size and it is proportionate to the Shadow hand model. We chose a set of ten objects, following [1]: cracker box, sugar box, mustard bottle, potted meat can, banana, pitcher base, bleach cleanser, mug, power drill, scissors.

In order to simplify the data acquisition, each object appears in a reachable position for the hand. In this way, the MPO agent can more easily learn a policy that maximizes the contact points between the fingertips and the different objects, without spending too much time to find the object in the scene first. We explicitly chose to keep the setup as simple as possible, to keep the focus of this work on the touch representations learned, rather than on the policy used to explore the space.

We vary the rotation of each object around its vertical axis, and the tilt of the object from the vertical axis. Rotations are sampled from the set of $\{0, \pm30, \pm60, \pm90, \pm120, \pm150\}$ degree angles. Tilt angles are sampled from the set $\{0, \pm15, \pm30\}$ degrees.

Two disjoint subsets are generated also for YCB objects to perform the evaluation on unseen objects. The two sets are as follows: Set1 = [banana, bleach, cracker box, driller, mug], Set2 = [mustard bottle, pitcher, potted meat, scissors, sugar box]. There are on average 10000 instances of each objects, presented with different orientations.

## 3 Tactile sensors

In simulation, each fingertip has a spatial touch sensor attached, with a spatial resolution of $4 \times 4$ and three channels: one for normal force and two for tangential forces (in Newtons). Thus, each fingertip has 16 channels that can collide with other objects, and that can sense 3D forces during such contacts. We simplify this by taking the absolute value and then summing across the spatial dimensions, to obtain a single 3D force vector for each fingertip.

On our real Shadow Hand, fingertips are equipped with BioTac® sensors [2, 3], which provide a more complex array of tactile signals. Despite the difference with the simulated sensors, readings from the pressure channel of each tactile sensor can be acquired and then normalized to match the range of the simulated tactile sensors. In this way, it is possible to directly compare results in simulation and on the real robot, without having to change the parameters of the task or learning algorithm. On the other hand, these adjustments are not necessary, since our proposed learning methods are totally agnostic and independent of the input dimensions and characteristics. Hence, new representations can be easily trained from real world robot data.

## 4 Network and training implementation details

We train the self-supervised cross-modal representation for $300\,000$ iterations with a batch size of 16 where each sample has 200 time steps. We finetuned the model for few-shot classification for $50\,000$ steps. We perform gradient-descent on GPU with an Adam optimizer using a fixed learning rate of $0.001$ for CM-GEN and $0.0001$ for all the other networks. The learning rate for few-shot finetuning is fixed to $0.0001$ for all feature types.

**Vision encoding**: $64 \times 64 \times 3$ frames are processed through a convolutional encoder with residual connections; we use 4 residual blocks with $[8, 16, 32, 64]$ channels and ReLU activations. Each residual block has 3 conv2D layers with $3 \times 3$ filters and $1 \times 1$ stride, followed by $3 \times 3$ pooling layers with $2 \times 2$ stride. The resulting features are flattened and downsized to 128 dimensions through a linear layer.

**Touch and proprioception encoding**: to process the concatenation of proprioception and touch we use a four-layers MLP with $[256, 256, 256, 128]$ channels respectively, and ReLUs nonlinearities. The MLP embeddings are processed by an LSTM with size 128, and skip-connections.

**$L^3$-net**: On top of the encoded multimodal pair 2-layer MLP with 256 channels (hidden layer size) and a linear layer for binary classification output are used. ReLU activations are applied in the MLP.

**CMDC**: Similar to $L^3$-net, 2-layer MLP with 256 channels with ReLU activations are utilized here. A final linear layer computes the logits for the multi-class classification. The predicted distance intervals are $\{[0], [1], [2-3], [4-5], [6-200]\}$.

**TCN and CMC**: $n$-pairs implementation doesn't require any extra layers.

**CM-CPC and CM-CPC-H**: For each time step a linear layer with embedding size of 128 is used to predict future (or history). CM-CPC is trained to predict 20 steps ahead. CM-CPC-H is trained to predict 20 steps into the future, the current frame, and 20 steps back into the history.

**CM-GEN**: The decoder network has 5 conv2D layers with $[256, 128, 64, 32, 16]$ channels and $3 \times 3$ filters. A $2 \times 2$ bilinear resize operator followed by leaky ReLU is applied after each convolutional layer. Finally a single conv2D layer with $3 \times 3$ filter is utilized to generate the $64 \times 64 \times 3$ image.

**Few-shot classifier**: This classifier is only utilized in few-shot finetuning stage. It is a 3-layers MLP, with size $[256, 256, c]$, where $c$ is the number of classes for the relevant classification task.

## 5 Supplementary experimental results

We report here other visualizations of our experimental results. The following tables show the proposed methodology performance on few-shot classification of unseen objects. Table 1 and 2 show results of few-shot object classification on unseen objects that belong to the same dataset. Table 3 shows the results of few-shot classification across the two datasets. Learned features are trained on one of the sets, and evaluated on few-shot classification on the other set, using 1, 3, 5, 10, 25, 50 or 100 examples for each object's class.

Fig. 1 reports all few-shot classification experiments on pose estimation on the YCB objects dataset. Table 4 reports all accuracy scores for all objects and all methods for the 10-shots experiment.

Table 1: Evaluation on unseen objects - Props classification, from Set2 to Set1, and from Set2 to Set1. The mean across these results is also reported. Darker colors indicate better scores.

|  | Props: set1 → set2 | | | | | | | | Props: set2 → set1 | | | | | | | |
|---|---|---|---|---|---|---|---|---|---|---|---|---|---|---|---|---|
|  | 1 | 3 | 5 | 10 | 25 | 50 | 100 | mean | 1 | 3 | 5 | 10 | 25 | 50 | 100 | mean |
| CM-CPC | 30.0% | 41.7% | 58.7% | 71.7% | 81.7% | 87.8% | 91.6% | 66.1% | 24.8% | 42.6% | 49.1% | 67.9% | 81.8% | 86.8% | 93.1% | 63.7% |
| CM-CPC-H | 29.5% | 46.0% | 60.3% | 73.1% | 83.1% | 86.9% | 92.6% | 67.4% | 22.1% | 41.1% | 60.2% | 71.7% | 83.9% | 88.8% | 92.6% | 65.8% |
| CM-GEN | 34.5% | 49.8% | 61.6% | 72.1% | 78.4% | 85.6% | 88.0% | 67.1% | 33.9% | 51.5% | 65.4% | 70.4% | 80.4% | 85.0% | 90.5% | 68.1% |
| CMC | 28.9% | 44.8% | 54.1% | 67.3% | 84.3% | 86.9% | 92.6% | 65.6% | 24.8% | 41.7% | 55.9% | 69.6% | 83.5% | 90.2% | 91.5% | 65.3% |
| CMDC | 26.9% | 46.9% | 57.0% | 71.8% | 80.5% | 90.7% | 92.8% | 66.7% | 28.6% | 41.4% | 55.5% | 70.8% | 81.3% | 89.3% | 92.4% | 65.6% |
| L3-NET | 27.2% | 38.9% | 52.4% | 70.7% | 84.7% | 87.4% | 91.8% | 64.7% | 26.1% | 40.9% | 52.1% | 68.5% | 80.8% | 88.1% | 93.8% | 64.3% |
| TCN | 26.7% | 43.8% | 56.9% | 70.3% | 82.6% | 88.1% | 92.5% | 65.8% | 28.6% | 44.8% | 56.9% | 69.0% | 85.8% | 89.1% | 91.6% | 66.5% |
| scratch | 29.2% | 41.1% | 57.3% | 71.2% | 82.3% | 90.7% | 90.2% | 66.0% | 21.6% | 40.4% | 56.8% | 66.8% | 79.4% | 85.0% | 89.1% | 62.7% |

Table 2: Evaluation on unseen objects - YCB objects classification, from Set1 to Set2, and from Set2 to Set1. The mean across these results is also reported. Darker colors indicate better scores.

|  | YCB: set1 → set2 | | | | | | | | YCB: set2 → set1 | | | | | | | |
|---|---|---|---|---|---|---|---|---|---|---|---|---|---|---|---|---|
|  | 1 | 3 | 5 | 10 | 25 | 50 | 100 | mean | 1 | 3 | 5 | 10 | 25 | 50 | 100 | mean |
| CM-CPC | 21.7% | 32.5% | 36.5% | 45.1% | 55.3% | 64.3% | 71.6% | 46.7% | 26.5% | 32.1% | 35.1% | 38.7% | 54.7% | 62.6% | 69.5% | 45.6% |
| CM-CPC-H | 29.0% | 37.1% | 36.5% | 46.7% | 58.5% | 67.0% | 71.2% | 49.4% | 29.8% | 32.3% | 37.6% | 45.7% | 60.7% | 63.6% | 73.1% | 49.0% |
| CM-GEN | 32.0% | 46.2% | 49.9% | 52.7% | 58.5% | 62.3% | 66.3% | 52.6% | 30.6% | 41.2% | 49.6% | 52.9% | 56.7% | 62.2% | 67.2% | 51.5% |
| CMC | 27.8% | 30.6% | 38.0% | 41.3% | 59.0% | 68.8% | 75.5% | 48.7% | 30.8% | 28.3% | 37.3% | 41.5% | 55.9% | 70.2% | 78.4% | 48.9% |
| CMDC | 29.2% | 31.4% | 39.1% | 40.5% | 60.4% | 69.1% | 77.5% | 49.6% | 29.3% | 29.5% | 39.5% | 41.9% | 59.1% | 67.3% | 74.8% | 48.8% |
| L3-NET | 27.5% | 29.6% | 33.1% | 42.7% | 57.5% | 69.0% | 75.8% | 47.9% | 29.8% | 36.7% | 36.0% | 43.5% | 57.5% | 67.0% | 76.6% | 49.6% |
| TCN | 30.1% | 33.5% | 34.0% | 43.0% | 57.2% | 69.4% | 73.6% | 48.7% | 31.7% | 30.4% | 37.2% | 37.7% | 56.2% | 67.2% | 76.5% | 48.1% |
| scratch | 26.2% | 28.9% | 33.6% | 40.3% | 53.1% | 51.3% | 59.9% | 41.9% | 24.5% | 33.6% | 32.2% | 41.0% | 50.6% | 62.5% | 61.8% | 43.7% |

Table 3: Evaluation on unseen objects across datasets, from Props to YCB objects, and from YCB objects to Props. he mean across these results is also reported. Darker colors indicate better scores.

|  | Props: all → YCB: all | | | | | | | | YCB: all → Props: all | | | | | | | |
|---|---|---|---|---|---|---|---|---|---|---|---|---|---|---|---|---|
|  | 1 | 3 | 5 | 10 | 25 | 50 | 100 | mean | 1 | 3 | 5 | 10 | 25 | 50 | 100 | mean |
| CM-CPC | 12.9% | 15.5% | 19.5% | 23.5% | 31.6% | 45.7% | 57.3% | 29.4% | 09.3% | 23.3% | 31.2% | 49.6% | 70.2% | 83.4% | 85.4% | 50.4% |
| CM-CPC-H | 14.0% | 16.1% | 16.6% | 26.8% | 33.4% | 44.7% | 59.6% | 30.2% | 10.0% | 25.3% | 31.6% | 56.2% | 74.8% | 85.5% | 86.9% | 52.9% |
| CM-GEN | 13.3% | 17.8% | 18.5% | 23.0% | 24.9% | 33.0% | 38.0% | 24.1% | 06.5% | 11.1% | 17.8% | 28.6% | 51.1% | 64.4% | 77.8% | 36.8% |
| CMC | 15.1% | 19.0% | 20.5% | 28.2% | 35.4% | 46.7% | 59.0% | 32.0% | 09.2% | 22.5% | 30.2% | 49.1% | 69.2% | 76.2% | 84.2% | 48.7% |
| CMDC | 15.4% | 16.8% | 21.3% | 29.0% | 37.0% | 47.6% | 58.2% | 32.2% | 10.4% | 23.8% | 30.1% | 49.1% | 72.4% | 78.1% | 85.7% | 49.9% |
| L3-NET | 15.9% | 17.4% | 18.9% | 25.9% | 37.1% | 47.6% | 57.8% | 31.5% | 11.5% | 22.2% | 26.8% | 46.5% | 72.0% | 82.6% | 86.6% | 49.7% |
| TCN | 16.4% | 19.7% | 21.8% | 26.9% | 36.5% | 46.2% | 57.7% | 32.2% | 11.2% | 24.8% | 31.6% | 50.5% | 72.9% | 83.2% | 85.4% | 51.4% |
| scratch | 13.1% | 15.2% | 17.9% | 25.0% | 30.1% | 36.0% | 48.7% | 26.6% | 09.2% | 23.9% | 28.1% | 48.6% | 64.5% | 79.2% | 84.1% | 48.2% |

3

Table 4: Evaluation on 10-shots pose estimation of all ten YCB objects. Darker colors indicate better scores.

|          | banana | bleach | cracker_box | driller | mug   | mustard_bottle | pitcher | potted_meat | scissors | sugar_box |
|----------|--------|--------|-------------|---------|-------|----------------|---------|-------------|----------|-----------|
| CM-CPC   | 37.3%  | 38.2%  | 23.4%       | 28.7%   | 43.0% | 19.7%          | 49.6%   | 36.6%       | 55.5%    | 18.6%     |
| CM-CPC-H | 42.7%  | 38.2%  | 25.5%       | 33.7%   | 46.3% | 20.1%          | 51.2%   | 42.6%       | 60.3%    | 22.3%     |
| CM-GEN   | 27.5%  | 25.7%  | 19.0%       | 25.9%   | 35.2% | 13.1%          | 33.9%   | 26.4%       | 49.1%    | 09.5%     |
| CMC      | 40.7%  | 38.0%  | 25.3%       | 28.7%   | 43.9% | 18.5%          | 52.0%   | 35.7%       | 59.1%    | 16.4%     |
| CMDC     | 38.4%  | 40.8%  | 28.1%       | 29.4%   | 44.6% | 20.1%          | 49.5%   | 37.4%       | 58.7%    | 19.8%     |
| L3-NET   | 41.4%  | 40.5%  | 28.6%       | 32.1%   | 41.2% | 20.2%          | 54.3%   | 38.4%       | 61.0%    | 18.1%     |
| TCN      | 43.8%  | 40.6%  | 25.8%       | 33.3%   | 50.0% | 18.1%          | 54.0%   | 35.2%       | 60.4%    | 16.5%     |
| scratch  | 38.7%  | 33.1%  | 20.8%       | 18.0%   | 41.7% | 17.1%          | 48.9%   | 34.8%       | 49.2%    | 11.5%     |

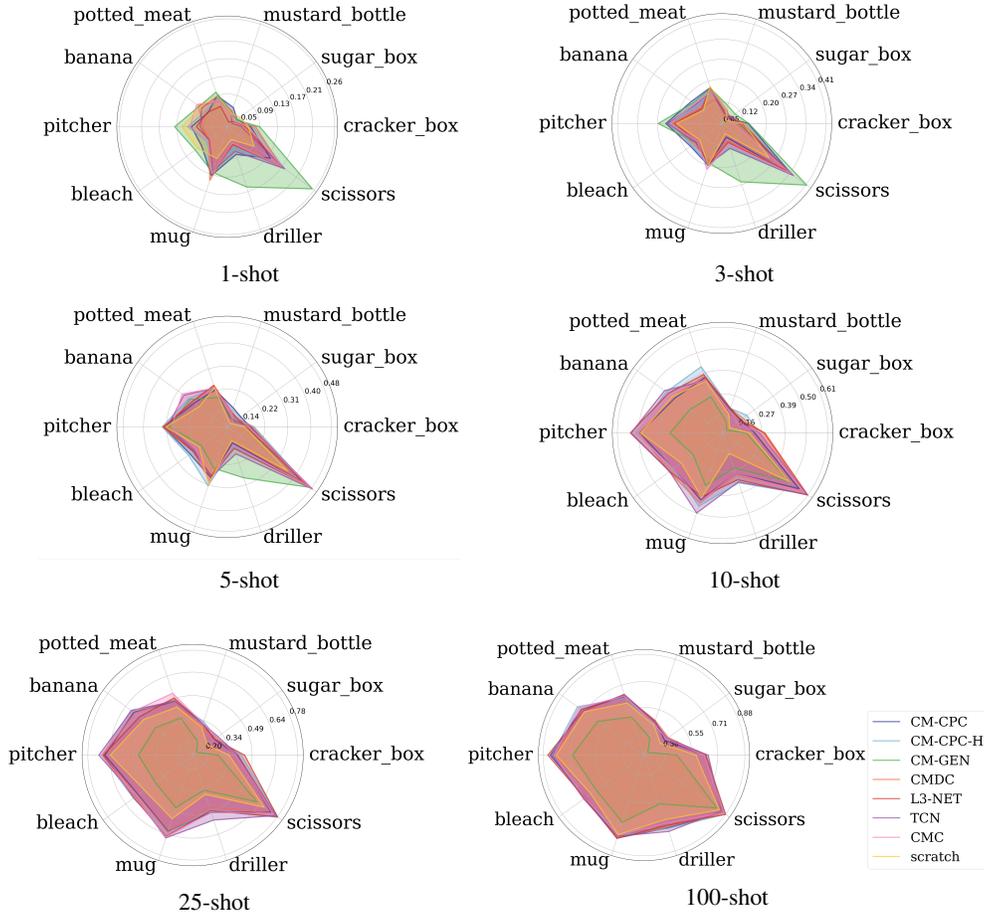

Figure 1: Pose estimation accuracy of YCB objects for {1, 3, 5, 10, 25, 50, 100}-shot classification. Larger area is better.



\end{document}